\documentclass{nle}
\usepackage{graphicx}
\let\bibhang\relax
\usepackage{natbib}
 \bibpunct[, ]{(}{)}{,}{a}{}{,}%
 \def\bibhang{24pt}%
\usepackage{xcolor}
\usepackage{url}
\usepackage{multirow}

\usepackage{arabtex}
\usepackage{utf8}
\setcode{utf8}

\title[Arabic Diacritic Recovery Using a Feature-Rich biLSTM Model]{Arabic Diacritic Recovery Using a Feature-Rich biLSTM Model}
\author[]{
    Kareem Darwish, Ahmed Abdelali, Hamdy Mubarak, Mohamed Eldesouki \\
    Qatar Computing Research Institute. Hamad Bin Khalifa University, Doha. Qatar\\
    \{kdarwish,aabdelali,hmubarak,mohamohamed\}@hbku.edu.qa
    }

\begin{document}
\label{firstpage}
\maketitle

\begin{abstract}
Diacritics (short vowels) are typically omitted when writing Arabic text, and readers have to reintroduce them to correctly pronounce words.  There are two types of Arabic diacritics: the first are core-word diacritics (CW), which specify the lexical selection, and the second are case endings (CE), which typically appear at the end of the word stem and generally specify their syntactic roles.  Recovering CEs is relatively harder than recovering core-word diacritics due to inter-word dependencies, which are often distant.  In this paper, we use a feature-rich recurrent neural network model that uses a variety of linguistic and surface-level features to recover both core word diacritics and case endings. Our model surpasses all previous state-of-the-art systems with a CW error rate (CWER) of 2.86\% and a CE error rate (CEER) of 3.7\% for Modern Standard Arabic (MSA) and CWER of 2.2\% and CEER of 2.5\% for Classical Arabic (CA). 
When combining diacritized word cores with case endings, the resultant word error rate is 6.0\% and 4.3\% for MSA and CA respectively. This highlights the effectiveness of feature engineering for such deep neural models.

\end{abstract}

\section{Introduction}

Modern Standard Arabic (MSA) and Classical Arabic (CA) have two types of vowels, namely long vowels, which are explicitly written, and short vowels, aka diacritics, which are typically omitted in writing but are reintroduced by readers to properly pronounce words. Since diacritics disambiguate the sense of the words in context and their syntactic roles in sentences, automatic diacritic recovery is essential for applications such as text-to-speech and educational tools for language learners, who may not know how to properly verbalize words. Diacritics have two types, namely: core-word (CW) diacritics, which are internal to words and specify lexical selection; and case-endings (CE), which appear on the last letter of word stems, typically specifying their syntactic role. For example, the word ``ktb''\footnote{Buckwalter encoding is used in this paper \cite{buckwalter2002buckwalter}} (\<كتب>) can have multiple diacritized forms such as ``k\textbf{a}t\textbf{a}b'' (\<كَتَب> -- meaning ``he wrote'') ``k\textbf{u}t\textbf{u}b'' (\<كُتُب> -- ``books'').  While ``katab'' can only assume one CE, namely ``fatHa'' (``a''), ``kutub'' can accept the CEs: ``damma'' (``u'') (nominal -- ex. subject), ``a'' (accusative -- ex. object), ``kasra'' (``i'') (genitive -- ex. PP predicate), or their nunations. 
There are 14 diacritic combinations. When used as CEs, they typically convey specific syntactic information, namely: \textbf{fatHa} ``a'' for accusative nouns, past verbs and subjunctive present verbs; \textbf{kasra} ``i'' for genitive nouns; \textbf{damma} ``u'' for nominative nouns and indicative present verbs; \textbf{sukun} ``o'' for jussive present verbs and imperative verbs. FatHa, kasra and damma can be preceded by \textbf{shadda} ``$\sim$'' for gemination (consonant doubling) and/or converted to \textbf{nunation} forms following some grammar rules. In addition, according to Arabic orthography and phonology, some words take a \textbf{virtual} (null) ``\#'' marker when they end with certain characters (ex: long vowels). This applies also to all non-Arabic words (ex: punctuation, digits, Latin words, etc.). Generally, function words, adverbs and foreign named entities (NEs) have set CEs (sukun, fatHa or virtual). 

Similar to other Semitic languages, Arabic allows flexible Verb-Subject-Object as well as Verb-Object-Subject constructs~\citep{Attia:08}. Such flexibility creates inherent ambiguity, which is resolved by diacritics as in 
``r$>$Y Emr Ely'' (\<رأى عمر علي> Omar saw Ali/Ali saw Omar). In the absence of diacritics it is not clear who saw whom. Similarly, in the sub-sentence ``kAn Alm\&tmr \textbf{AltAsE}'' (\<كان المؤتمر التاسع>), 
if the last word, is a predicate of the verb ``kAn'', then the sentence would mean ``this conference was the ninth'' and would receive a fatHa (a) as a case ending.  Conversely, if it was an adjective to the ``conference'', then the sentence would mean ``the ninth conference was ...'' and would receive a damma (u) as a case ending.  Thus, a consideration of context is required for proper disambiguation. Due to the inter-word dependence of CEs, they are typically harder to predict compared to core-word diacritics~\citep{habash2007arabic,roth2008arabic,harrat:hal-00925815,ameur2015restoration}, with CEER of state-of-the-art systems being in double digits compared to nearly 3\% for word-cores. Since recovering CEs is akin to shallow parsing~\citep{marton2010improving} and requires morphological and syntactic processing, it is a difficult problem in Arabic NLP. In this paper, we focus on recovering both CW diacritics and CEs. We employ two separate Deep Neural Network (DNN) 
architectures for recovering both kinds of diacritic types.  We use character-level and word-level bidirectional Long-Short Term Memory (biLSTM) based recurrent neural models for CW diacritic and CE recovery respectively.  We train models for both Modern Standard Arabic (MSA) and Classical Arabic (CA).  For CW diacritics, the model is informed using word segmentation information and a unigram language model. We also employ a unigram language model to perform post correction on the model output. We achieve word error rates for CW diacritics of 2.9\% and 2.2\% for MSA and CA.  The MSA word error rate is 6\% lower than the best results in the literature (the RDI diacritizer \citep{rashwan2015deep}).  The CE model is trained with a rich set of surface, morphological, and syntactic features. 
The proposed features would aid the biLSTM model in capturing syntactic dependencies 
indicated by Part-Of-Speech (POS) tags, gender and number features, morphological patterns, and affixes.  
We show that our model achieves a case ending error rate (CEER) of 3.7\% for MSA and 2.5\% for CA.  For MSA, this CEER is more than 60\% lower than other state-of-the-art systems such as Farasa and the RDI diacritizer, which are trained on the same dataset and achieve CEERs of 10.7\% and 14.4\% respectively. 
The contributions of this paper are as follows:
\begin{itemize}
\item We employ a character-level RNN model that is informed using word morphological information and a word unigram language model to recover CW diacritics.  Our model beats the best state-of-the-art system by 6\% for MSA.
\item We introduce a new feature rich RNN-based CE recovery model that achieves errors rates that are 60\% lower than the current state-of-the-art for MSA.
\item We explore the effect of different features, which may potentially be exploited for Arabic parsing.
\item We show the effectiveness of our approach for both MSA and CA.
\end{itemize}

\section{Background}
\label{sec:background}

Automatic diacritics restoration has been investigated for many different language such as European languages (e.g. Romanian~\citep{mihalcea2002diacritics,tufics2008diac+}, French \citep{zweigenbaum2002restoring}, and Croatian \citep{vsantic2009automatic}), African languages (e.g. Yorba \citep{orife2018attentive}), Southeast Asian languages (e.g. Vietnamese \citep{luu2012pointwise}), Semitic language (e.g. Arabic and Hebrew \citep{gal2002hmm}), and many others \citep{de2007automatic}.   For many languages, diacritic (or accent restoration) is limited to a handful of letters.  However, for Semitic languages, diacritic recovery extends to most letters. Many general approaches have been explored for this problem including linguistically motivated rule-based approaches, machine learning approaches, such as Hidden Markov Models (HMM) \citep{gal2002hmm} and Conditional Random Fields (CRF) \citep{darwish2018diacritization}, and lately deep learning approaches 
such as Arabic \citep{abandah2015automatic,Hifny2018Hybrid,mubarak2019highly}, Slovak \citep{hucko2018Diacritic}, and Yorba \citep{orife2018attentive}.  

Aside from rule-based approaches  
~\citep{el1989arabic}, different methods were used to recover diacritics in Arabic text. Using a hidden Markov model (HMM) \citep{gal2002hmm,elshafei2006statistical} with an input character sequence, the model attempts to find the best state sequence given previous observations.  \cite{gal2002hmm} reported a 14\% word error rate (WER) while \cite{elshafei2006statistical} achieved a 4.1\% diacritic error rate (DER) on the Quran (CA). \cite{vergyri2004automatic} combined both morphological, acoustic, and contextual features to build a diacritizer trained on FBIS and LDC CallHome ECA collections. They reported a 9\% (DER) without CE, and 28\% DER with CE. \cite{nelken2005arabic} employed a cascade of a finite state transducers. The cascade stacked a word language model (LM), a charachter 
LM, and a morphological model. The model achieved an accuracy of 7.33\% WER without CE and and 23.61\% WER with CE. \cite{zitouni2006maximum} employed a maximum entropy model for sequence classification. The system was trained on the LDC’s Arabic Treebank (ATB) 
and evaluated on a 600 articles from An-Nahar Newspaper (340K words) and achieved 5.5\% DER and 18\% WER on words without CE.\\ 
\cite{bebah2014hybrid} used a hybrid approach that utilizes the output of Alkhalil morphological Analyzer \citep{bebah2011alkhalil} to generate all possible out of context diacritizations of a word. Then, an HMM guesses the correct diacritized form. 
Similarly, Microsoft Arabic Toolkit Services (ATKS) diacritizer \citep{microsoft2013diac} uses a rule-based morphological analyzer that produces possible analyses and an HMM in conjunction with rules to guess the most likely analysis.  They report WER of 11.4\% and 4.4\% with and without CE.
MADAMIRA ~\citep{pasha2014madamira} uses a combinations of morpho-syntactic features to rank a list of potential analyses provided by the Buckwalter Arabic Morphological Analyzer (BAMA) \citep{buckwalter2004buckwalter}. An SVM trained on ATB selects the most probable analysis, including the diacritized form. MADAMIRA achieves 19.0\% and 6.7\% WER with and without CE respectively \citep{darwish2017arabic}.
Farasa \citep{darwish2017arabic} uses an HMM to guess CW diacritics and an SVM-rank based model trained on morphological and syntactic features to guess CEs. Farasa achieves WER of 12.8\% and 3.3\% with and without CEs. 

More recent work employed different neural architectures to model the diacritization problem. 
\cite{abandah2015automatic} used a biLSTM recurrent neural network trained on the same dataset as \citep{zitouni2006maximum}. They explored one, two and three BiLSTM layers with 250 nodes in each layers, achieving WER of 9.1\% including CE on ATB. Similar architectures were used but achieved lower results \citep{rashwan2015deep,belinkov2015arabic}.
\cite{azmi2015survey} provide a comprehensive survey on Arabic diacritization. A more recent survey by \cite{HamedZesch:2017} concluded that reported results are often incomparable due to the usage of different test sets. 
They 
concluded that a large unigram LM for CW diacritic recovery is competitive with many of the systems in the literature, which prompted us to utilize a unigram language model for post correction. 
As mentioned earlier, two conclusions can be drawn, namely:  restoring CEs is more challenging than CW diacritic restoration; and combining multiple features typically improves CE restoration. 

In this paper, we expand upon the work in the literature by introducing feature-rich DNN models for restoring both CW and CE diacritics.  We compare our models to multiple systems on the same test set.  We achieve results that reduce diacritization error rates by more than half compared to the best SOTA systems. We further conduct an ablation study to determine the relative effect of the different features. 

As for Arabic, it is a Semitic language with derivational morphology. Arabic nouns, adjectives, adverbs, and verbs are typically derived from a closed set of 10,000 roots of length 3, 4, or rarely 5.  Arabic nouns and verbs are derived from roots by applying templates to the roots to generate stems.  Such templates may carry information that indicate morphological features of words such POS tag, gender, and number.  For example, given a 3-letter root with 3 consonants CCC, a valid template may be CwACC , where the infix ``wA'' (\<وا>) is inserted, this template typically indicates an Arabic broken, or irregular, plural template for a noun of template CACC or CACCp if masculine or feminine respectively. Further, stems may accept prefixes and/or suffixes to form words. Prefixes include coordinating conjunctions, determiner, and prepositions, and suffixes include attached pronouns and gender and number markers.

\section{Our Diacritizer}

\subsection{Training and Test Corpora}
\label{rdi-data}
For MSA, we acquired the diacritized corpus that was used to train the RDI \citep{rashwan2015deep} diacritizer and the Farasa diacritizer \citep{darwish2017arabic}.  The corpus contains 9.7M tokens with approximately 194K unique surface forms (excluding numbers and punctuation marks).  The corpus covers multiple genres such as politics and sports and is a mix of MSA and CA. This corpus is considerably larger than the Arabic Treebank~\citep{maamouri2004atb3} and is more consistent in its diacritization. For testing, we used the freely available WikiNews test set \citep{darwish2017arabic}, which is composed of 70 MSA WikiNews articles (18,300 tokens) and evenly covers a variety of genres including politics, economics, health, science and technology, sports, arts and culture.

For CA, we obtained a large collection of fully diacritized classical texts (2.7M tokens) from a book publisher, and we held-out a small subset of 5,000 sentences (approximately 400k words) for testing. Then, we used the remaining sentences to train the CA models. 

\subsection{Core Word Diacritization}

\subsubsection{Features.}
Arabic words are typically derived from a limited set of roots by fitting them into so-called stem-templates (producing stems) and may accept a variety of prefixes and suffixes such as prepositions, determiners, and pronouns (producing words). Word stems specify the lexical selection and are typically unaffected by the attached affixes. We used 4 feature types, namely:
\begin{itemize}
\item \textbf{CHAR:} the characters.
\item \textbf{SEG:} the position of the character in a word segment.  For example, given the word ``wAlktAb'' (\<والكتاب> and the book/writers), which is composed of 3 segments ``w+Al+ktAb'' (\<و+ال+كتاب>). Letters were marked as ``B'' if they begin a segment, ``M'' if they are in the middle of a segment, ``E'' if they end a segment, and ``S'' if they are single letter segments.  So for ``w+Al+ktAb'', the corresponding character positions are ``S+BE+BMME''.  We used Farasa to perform segmentation, which has a reported segmentation accuracy of 99\% on the WikiNews dataset \citep{DARWISH2016farasa}.  
\item \textbf{PRIOR:} diacritics seen in the training set per segment. Since we used a character level model, this feature informed the model with word level information. For example, the word ``ktAb'' (\<كتاب>) was observed to have two diacritized forms in the training set, namely ``k\textbf{i}t\textbf{a}Ab'' (\<كِتَاب> -- book) and ``k\textbf{u}t$\sim$\textbf{a}Ab'' (\<كُتَّاب> -- writers).  The first letter in the word (``k'') accepted the diacritics ``i'' and ``u''. Thus given a binary vector representing whether a character is allowed to assume any of the eight primitive Arabic diacritic marks (a, i, u, o, K, N, F, and $\sim$ in order), the first letter would be given the following vector ``01100000''. If a word segment was never observed during training, the vector for all letters therein would be set to 11111111.  This feature borrows information from HMM models, which have been fairly successful in diacritizing word cores.
\item \textbf{CASE:} whether the letter expects a core word diacritic or a case ending. Case endings are placed on only one letter in a word, which may or may not be the last letter in the word.  This is a binary feature.
\end{itemize}

\subsubsection{DNN Model.}
\begin{figure}
\includegraphics[width=0.7\linewidth]{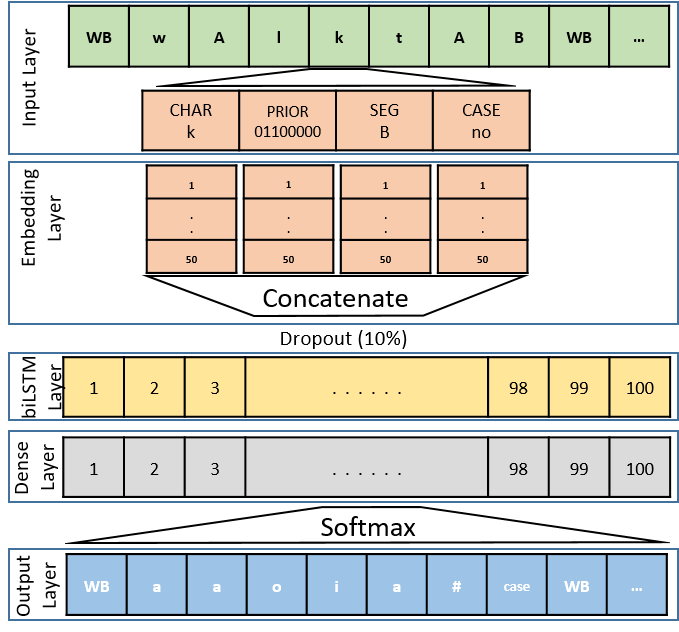}
\caption{DNN model for core word diacritics}
\label{fig:dnn-core}
\end{figure}

Using a DNN model, particularly with a biLSTM ~\citep{schuster1997bilstm}, is advantageous because the model automatically explores the space of feature combinations and is able to capture distant dependencies. A number of studies have explored various biLSTM architectures \citep{abandah2015automatic,rashwan2015deep,belinkov2015arabic} including stacked biLSTMs confirming their effectiveness. As shown in Figure \ref{fig:dnn-core}, we employed a character-based biLSTM model with associated features for each character. Every input character had an associated list of $m$ features, and we trained randomly initialized embeddings of size 50 for each feature.  Then, we concatenated the feature embeddings vectors creating an $m\times50$ vector for each character, which was fed into the biLSTM layer of length 100. The output of the biLSTM layer was fed directly into a dense layer of size 100. 
We used early stopping with patience of 5 epochs, a learning rate of 0.001, a batch size of 256, and an Adamax optimizer. 
The input was the character sequence in a sentence with words being separated by word boundary markers (WB), and we set the maximum sentence length to 1,250 characters.

\subsection{Case Ending Diacritization}

\subsubsection{Features.}
Table \ref{table:features} lists the features that we used for CE recovery.  We used Farasa to perform segmentation and POS tagging and to determine stem-templates \citep{darwish2017arabic}. Farasa has a reported POS accuracy of 96\% on the WikiNews dataset \cite{darwish2017arabic}. Though the Farasa diacritizer utilizes a combination of some the features presented herein, namely segmentation, POS tagging, and stem templates, Farasa's SVM-ranking approach requires 
explicit specification of feature combinations (ex. $Prob(CE\|current\_word, prev\_word, prev\_CE)$). Manual exploration of the feature space is undesirable, and ideally we would want our learning algorithm to do so automatically. The flexibility of the DNN model allowed us to include many more surface level features such as affixes, leading and trailing characters in words and stems, and the presence of words in large gazetteers of named entities. As we show later, these additional features significantly lowered CEER. 

\begin{table*}[hbt!]
\begin{center}
\begin{scriptsize}
\begin{tabular}{p{2.5cm}|p{2.9cm}|p{6.5cm}}
Feature & Example & Explanation and Motivation \\ \hline
word	& w+b+mktb+t+nA (\<و+ب+مكتب+ت+نا> -- and in our library) & Some words have a fixed set of observed CEs	\\ \hline
word POS & CONJ+PREP+NOUN +NSUFF+PRON & Some POS combinations allow a closed set of CEs \\ \hline
gender/number & feminine/singular & Gender/number agreement (dis)allow certain attachments and may allow/exclude certain CEs \\ \hline
stem & mktb+p (\<مكتب+ة> -- library) & We attach gender and number noun suffixes such the singular feminine marker ``p'' (\<ـة>) because 
CEs appear on them. \\ \hline
stem POS & NOUN+NSUFF & Same rationale as word POS \\ \hline
prefix(es) \& POS & w+b+ (\<و+ب+>) \& CONJ+PREP & Certain prefixes affect CE directly. For example, the PREP ``b+'' (\<ب+>) is a preposition causing their noun predicates to assume the genitive case \\ \hline
suffix(es) \& POS & ``+nA'' (\<+نا>) \& PRON & Certain suffixes affect CE directly \\ \hline
stem template & mfEl+p (\<مفعل+ة> -- derived from the root ``ktb'' \<كتب>) & Some stem templates allow certain CEs and exclude others. Ex. the stem template ``$>$fEl'' (\<أفعل>) disallows tanween (``N'', ``K'', ``F'') \\ \hline
word/stem head/tail char uni/bi-grams & word: w (\<و>), wb (\<وب>); stem: A (\<ا>), nA (\<نا>) & Such characters can capture some morphological and syntactic information. Ex. verbs in present tense typically start with ``$>$ (\<أ>), n (\<ن>), y (\<ي>), or t (\<ت>)''. \\ \hline
sukun word & foreign NEs: ex. jwn (\<جون> -- John) & CE of certain words is strictly \textit{sukun}. We built a list from training set.\\ \hline
named entities & NEs & Named entities are more likely to have \textit{sukun} as CE.  We extracted the named entity list from the Farasa named entity recognizer \citep{darwish2013named,darwish2014simple}.\\
\end{tabular}
\caption{Features with examples and motivation.}
\label{table:features}
\end{scriptsize}
\end{center}
\end{table*}


\subsubsection{DNN Model}
\begin{figure}
\begin{center}
\includegraphics[width=0.7\linewidth]{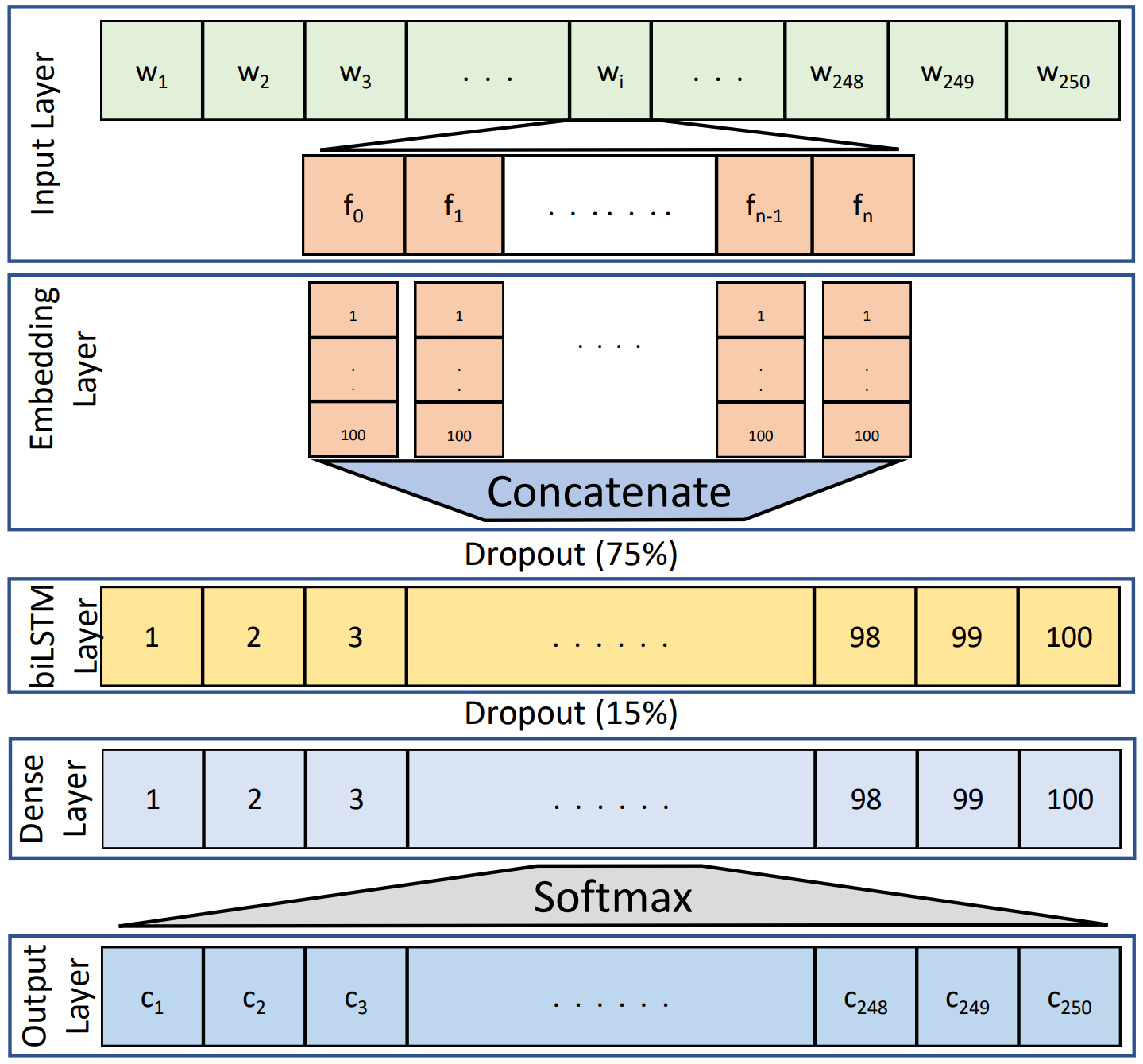}
\caption{DNN case ending model architecture}
\label{fig:dnn-model}
\end{center}
\end{figure}

Figure \ref{fig:dnn-model} shows the architecture of our DNN algorithm.  Every input word had an associated list of $n$ features, and we trained randomly initialized embeddings of size 100 for each feature.  Then, we concatenated the feature embeddings vectors creating an $n\times100$ vector for each word.  We fed these vectors into a biLSTM layer of 100 dimensions after applying a dropout of 75\%, where dropout behaves like a regularlizer to avoid overfitting~\citep{hinton2012improving}. We conducted side experiments with lower dropout rates, but the higher dropout rate worked best. The output of the biLSTM layer was fed into a 100 dimensional dense layer with 15\% dropout and softmax activation.  We conducted side experiments where we added additional biLSTM layers and replaced softmax with a conditional random field layer, but we did not observe improvements.  Thus, we opted for a simpler model. We used a validation set to determine optimal parameters such as dropout rate. 
Again, we used the ``Adamax'' optimizer with categorical cross entropy loss and a learning rate of 0.001. We also applied early stopping with patience of up to 5 consecutive epochs without improvement.  

\section{Experiments and Results}
\subsection{Core Word}

\subsubsection{Experimental Setup}
For all the experiments conducted herein, we used the Keras toolkit \citep{chollet2015keras} with a TensorFlow backend \citep{tensorflow2015-whitepaper}. We used the entirety of the training set as input, and we instructed Keras to use 5\% of the data for tuning (validation). 
We included the CASE feature, which specifies whether the letter accepts a normal diacritic or case ending, in all our setups. We conducted multiple experiment using different features, namely:
\begin{itemize}
\item \textbf{CHAR}: This is our baseline setup where we only used the characters as features.
\item \textbf{CHAR+SEG}: This takes the characters and their segmentation information as features.
\item \textbf{CHAR+PRIOR}: This takes the characters and their the observed diacritized forms in the training set.
\item \textbf{All}: This setup includes all the features.
\end{itemize}
We also optionally employed post correction.  For words that were seen in training, if the model produced a diacritized form that was not seen in the training data, we assumed it was an error and replaced it with the most frequently observed diacritized form (using a unigram language model).  We report two error rates, namely WER (at word level) and DER (at character level).  
We used relaxed scoring where we assumed an empty case to be equivalent to \textit{sukun}, and we removed default diacritics -- \textit{fatHa} followed by \textit{alef}, \textit{kasra} followed by \textit{ya}, and \textit{damma} followed by \textit{wa}. Using such scoring would allow to compare to other systems in the literature that may use different diacritization conventions.

\subsubsection{Results and Error analysis} For testing, we used the aforementioned WikiNews dataset to test the MSA diacritizer and the held-out 5,000 sentences for CA. Table \ref{res:coreWord} shows WER and DER results using different features with and without post correction. 

\begin{table}[hbt!]
\begin{center}
\begin{tabular}{l|c|c|c|c||c|c|c|c}
        & \multicolumn{4}{c|}{MSA}   &   \multicolumn{4}{c}{CA} \\ \hline
 		&	\multicolumn{2}{c|}{DNN} &	\multicolumn{2}{c|}{DNN+Post} &	\multicolumn{2}{c|}{DNN} &	\multicolumn{2}{c}{DNN+Post}\\ \hline
Model	&	WER	&	DER	&	WER	&	DER 	&	WER	&	DER 	&	WER	&	DER	\\ \hline
CHAR	&	3.5	&	1.1	&	3.3	&	1.0	 & 5.1 & 2.1 & 2.7 & 1.0 \\
CHAR+SEG	&	3.3	&	1.1	&	3.2	&	1.0	 & 4.7 & 1.9 & 2.6 & 1.0 \\
CHAR+PRIOR	&	3.8	&	1.2	&	3.7	&	1.1	& 3.8 & 1.6 & 2.3 & 0.9 \\ \hline
\textbf{ALL}	&	\textbf{3.0}	&	\textbf{1.0}	&	\textbf{2.9}	&	\textbf{0.9} & \textbf{3.6} & \textbf{1.5}	& \textbf{2.2} & \textbf{0.9} \\
\end{tabular}
\caption{Core word diacritization results}
\label{res:coreWord}
\end{center}
\end{table}

\paragraph{\textbf{MSA Results:}} For MSA, though the CHAR+PRIOR feature led to worse results than using CHAR alone, the results show that combining all the features achieved the best results. 
Moreover, post correction improved results overall. 
We compare our results to five other systems, namely Farasa~\citep{darwish2017arabic},
MADAMIRA~\citep{pasha2014madamira}, 
RDI (Rashwan et al., 2015), MIT (Belinkow and Glass, 2015), and Microsoft ATKS \citep{microsoft2013diac}.  Table \ref{res:coreWordComparison} compares our system with others in the aforementioned systems. As the results show, our results beat the current state-of-the-art.

For error analysis, we analyzed all the errors (527 errors).  The errors types along with examples of each are shown in Table \ref{table:CWErrorTypes}. The most prominent error type arises from the selection of a valid diacritized form that does not match the context (40.8\%). Perhaps, including POS tags as a feature or augmenting the PRIOR feature with POS tag information and a bigram language model may reduce the error rate further.  The second most common error is due to transliterated foreign words including foreign named entities (23.5\%).  Such words were not observed during training.  Further, Arabic Named entities account for 10.6\% of the errors, where they were either not seen in training or they share identical non-diacritized forms with other words.  Perhaps, building larger gazetteers of diacritized named entities may resolve NE related errors.  In 10.8\% of the cases, the diacritizer produced in completely incorrect diacritized forms. In some the cases (9.1\%), though the diacritizer produced a form that is different from the reference, both forms were in fact correct.  Most of these cases were due to variations in diacritization conventions (ex. ``bare alef'' (A) at start of a word receiving a diacritic or not).  Other cases include foreign words and some words where both diacritized forms are equally valid.

\begin{table*}[hbt!]
\begin{center}
\begin{tabular}{p{2.2cm}|r|r|p{3.8cm}|p{3.8cm}}

Error	&	Freq.	&	\% & Explanation & Examples	\\	\hline
Wrong selection	&	215	&	40.8  & Homographs with different diacritized forms & ``qaSor'' (\<قَصْر> -- palace) vs. ``qaSar'' (\<قَصَر> -- he limited) \\ \hline
Foreign word	&	124	&	23.5 & transliterated words including 96 foreign named entities & wiykiymaAnoyaA (\<وِيكِيمَانْيَا>	-- Wikimania)  \\	\hline
Invalid diacritized form	&	57	&	10.8 & invalid form &  ya*okur (\<يّذْكُر>	-- he mentions) vs. ya*okar (\<يّذْكَر>) \\ \hline	
Named entity	&	56	&	10.6 & Arabic named entities & ``Eab\~Adiy'' (\<عَبَّادِي> -- name) vs. ``EibAdiy'' (\<عِبَادِي> -- my servants) \\ \hline
both correct	&	48	&	9.1	& Some words have multiple valid diacritized forms & ``wikAlap'' (\<وِكَالَة>) and ``wakAlap'' (\<وَكَالَة> -- agency) \\ \hline	
Affix diacritization error	&	16	&	3.0	& Some sufixes are erroneously diacritized & b\textbf{\textit{a}}Akt\$Afihim (\<بَاكتشافِهِم> -- with their discovery) \\ \hline
Reference is wrong	&	10	&	1.9	& the truth diacritics were incorrect & AlofiyfaA (\<الْفِيفَا>	-- FIFA) vs. AlofayofaA (\<الْفَيْفَا>) \\ \hline
dialectal word	&	1	&	0.2	 & dialectal word & mawaAyiliy (\<مَوَايِلِي>	-- my chant) \\ \hline
\end{tabular}
\caption{Error analysis: Core word error types for MSA}
\label{table:CWErrorTypes}
\end{center}
\end{table*}

\begin{table}[hbt!]
    \centering
    \begin{tabular}{c|r|r|p{8cm}}
Error	&	Count	&	\% & Most Common Causes	\\ \hline
a	$\Leftrightarrow$	u 	&	133	&	19.3	& \textit{POS error}: ex. ``ka\$afa'' (\<كَشَفَ> -- he exposed) vs. ``ka\$ofu'' (\<كَشْفُ> -- exposure) \& \textit{Subject vs. object}: ex. ``tuwHy \textbf{mivolu}'' (\<تُوحِي مِثْلُ> -- \textbf{such} indicates) vs. ``tuwHy \textbf{mivola}'' (\<تُوحِي مِثْلَ> -- she indicates \textbf{such}) \\
i	$\Leftrightarrow$	a	&	130	&	18.9	& \textit{Incorrect attachment} (due to coordinating conjunction or distant attachment): ex. ``Alogaza Alomusay\textasciitilde{}ili lilidumuEi -- \textbf{wa+AlraSaSi} vs. \textbf{wa+AlraSaSa} (\<الغَازَ الْمُسَيِّلَ لِلدُمُوعِ والرَصَاص> -- tear gas and bullets) where bullets were attached incorrectly to tear instead of gas \& \textit{indeclinability such as foreign words and feminine names}: ex. ``kaAnuwni'' (\<كَانُونِ> -- Cyrillic month name) vs. ``kaAuwna'' (\<كَانُونَ>) \\
i	$\Leftrightarrow$	u	&	95	&	13.8 & \textit{POS error of previous word}: ex. ``tadahowuru \textbf{waDoEihi}'' (\<تَدَهْؤُرُ وَضْعِهِ> -- deterioration of his situation -- situtation is part of idafa construct) vs. ``tadahowara \textbf{waDoEihu}'' (\<تَذَهْوَرَ وَضْعُهُ> -- his situation deteriorated -- situation is subject)	
\& \textit{Incorrect attachment} (due to coordinating conjunction or distant attachment): (as example for i $\Leftrightarrow$ a)\\
a	$\Leftrightarrow$	o	&	60	&	8.7	& \textit{Foreign named entities}: ex. ``siyraAloyuna'' (\<سِيرَالْيُونَ> -- Siera Leon) vs. ``siyraAloyuno'' (\<سِيرَالْيُونْ>) \\
i	$\Leftrightarrow$	K	&	27	&	4.0	 & \textit{Incorrect Idafa}: ``\textbf{liAt\textasciitilde{}ifaqi} ha\*aA Alo$>$usobuwE'' (\<لِاتِفَاقِ هَذَا الأُسْبُوع> -- this week's agreement) vs. ``\textbf{liAt\textasciitilde{}ifaqK} ha\*aA Alo$>$usobuwE'' (\<لِاتِّفَاقٍ هَذَا الْأُسْبُوع> -- to an agreement this week)\\
K	$\Leftrightarrow$ N	&	29	&	4.2 & \textit{Subject vs. object} (as in a $\Leftrightarrow$ u)	 and Incorrect attachment (as in i	$\Leftrightarrow$ a) \\
F	$\Leftrightarrow$	N	&	25	&	3.7 & \textit{Words ending with feminine marker ``p'' or ``At''}: ex. ``muHaADarap'' (\<مُحَاضَرَة> -- lecture)	\\
i	$\Leftrightarrow$ o	&	22	&	3.2 & \textit{Foreign named entities} (as in a	$\Leftrightarrow$ o)	\\
F	$\Leftrightarrow$	a	&	16	&	2.3	& \textit{Incorrect Idafa} (as in i	$\Leftrightarrow$ K) \\
u	$\Leftrightarrow$ o	&	14	&	2.0 & 	\textit{Foreign named entities} (as in a	$\Leftrightarrow$ o) \\
F	$\Leftrightarrow$	K	&	9	&	1.3 & 	\textit{Words ending with feminine marker} (as in F	$\Leftrightarrow$ N)\\
K	$\Leftrightarrow$	a	&	8	&	1.2	&  \textit{Incorrect Idafa} (as in i	$\Leftrightarrow$ K) \\
    \end{tabular}
    \caption{MSA case errors accounting from more than 1\% of errors}
    \label{tab:msaCaseErrors}
\end{table}

\begin{table}[hbt!]
    \centering
    \begin{tabular}{c|r|r|p{0.7\textwidth}}
Error	&	Count	&	\% & Most Common Causes	\\ \hline
a $\Leftrightarrow$ u & 2,907 & 28.4 & \textit{Subject vs. object}: ex. ``wafaqa \textbf{yawoma}'' (\<وَفَقَ يَوْمَ> -- he matches the day) vs. ex. ``wafaqa \textbf{yawomu}'' (\<وَفَقَ يَوْمُ> -- the day matches) \&  \textit{False subject} (object behaves like subject in passive tense): ex. ``yufar\textasciitilde{}iqu \textbf{qaDaA'a}'' (\<يُفَرِّقُ الْقَضَاءَ> -- he separates the make up) vs. ``yufar\textasciitilde{}aqu \textbf{qaDaA'u}'' (\<يُفَرَّقٌ الْقَضَاءُ> -- the make up is separated) \& 
\textit{Incorrect attachment} (due to coordinating conjunction): ex. ``f+a$>$aEohadu'' (\<فَأَعْهَدَ> -- so I entrust) vs. ``f+a$>$aEohadu'' (\<فَأَعْهِدُ>) \\
i $\Leftrightarrow$ u & 1,316 & 12.9 & \textit{Incorrect attachment} (due to coordinating conjunctions or distant attachment): (as in a $\Leftrightarrow$ u) \\
i $\Leftrightarrow$ a & 1,019 & 10.0 & \textit{Incorrect attachment} (as in a $\Leftrightarrow$ u) \& \textit{Indeclinability such as foreign words and feminine names}: ex. ``$>$ajoyaAdiyni'' (\<أَجْيَادِينِ> -- Ajyadeen (city name)) vs. ``$>$ajoyaAiyna'' (\<أَجْيَادِينَ>)
\\
a $\Leftrightarrow$ \# & 480 & 4.7 & \textit{Problem with reference where the case for some words, particularly non-Arabic names, is not provided in the reference:} ex. ``$<$isoHaAq'' (\<إِسْحَاق> -- Issac) vs. ``$<$isoHaAqa'' (\<إِسْحَاقَ>)\\
u $\Leftrightarrow$ \# & 426 & 4.2 & same problems as in a $\Leftrightarrow$ \#\\
K $\Leftrightarrow$ i & 371 & 3.6 & \textit{Incorrect Idafa}: ex. ``\textbf{EaTaA'i} Alofaqiyh'' (\<عَطَاءِ الْفَقِيه> -- the providence of the jurist) vs. ``\textbf{EaTaA'K} Alofaqiyh'' (\<عَطَاءٍ الْفَقِيه> -- Ataa the jurist) \\
K $\Leftrightarrow$ a & 328 & 3.2 & \textit{words ending with feminine marker}: ex. ``tayomiyap'' (\<تَيْمِيَة> --Taymiya)
\& \textit{Indeclinability:} ex. ``bi$<$i\$obiyliy\textasciitilde{}ap'' (\<وَبِإِشْبِيلِيَّة> -- and in Lisbon)\\
u $\Leftrightarrow$ o & 300 & 2.9 & \textit{confusion between past, present, and imperative moods of verbs and preceding markers (imperative ``laA'' vs. negation ``laA):} ex. ``laA tano\$ariHu'' (\<لا تَنْشَرِحُ> -- does not open up) vs. ``laA tano\$ariHo'' (\<لا تَنْشَرِحْ> -- do not open up)\\
a $\Leftrightarrow$ o & 278 & 2.7 & \textit{confusion between past, present, and imperative moods of verbs} (as in u $\Leftrightarrow$ o) \\
K $\Leftrightarrow$ N & 253 & 2.5 & \textit{Incorrect attachment} (as in i $\Rightarrow$ u)\\
N $\Leftrightarrow$ u & 254 & 2.5 & \textit{Incorrect Idafa} (as in K $\Rightarrow$ i)\\
F $\Leftrightarrow$ N & 235 & 2.3 & \textit{words ending with feminine marker} (as in K $\Rightarrow$ a)\\
i $\Leftrightarrow$ o & 195 & 1.9 & \textit{Differing conventions concerning handling two consecutive letters with sukun}: ex. ``\textbf{Eano} Aboni'' (\<عَنْ ابْنِ> -- on the authority of the son of)\ vs. ``\textbf{Eani} Aboni'' (\<عَنِ ابْنِ>) \\
i $\Leftrightarrow$ \# & 178 & 1.7 & same errors as for a $\Rightarrow$ \# \\
o $\Leftrightarrow$ \# & 143 & 1.4 & same errors as for a $\Rightarrow$ \# \\
    \end{tabular}
    \caption{CA case errors accounting from more than 1\% of errors}
    \label{tab:caCaseErrors}
\end{table}


\paragraph{\textbf{CA Results:}} For CA results, the CHAR+SEG and CHAR+PRIOR performed better than using characters alone with CHAR+PRIOR performing better than CHAR+SEG.  As in the case with MSA, combining all the features led to the best results.  Post correction had a significantly larger positive impact on results compared to what we observed for MSA.  This may indicate that we need a larger training set.  The best WER that we achieved for CW diacritics with post corrections is 2.2\%.  Since we did not have access to any publicly available system that is tuned for CA, we compared our best system to using our best MSA system to diacritize the CA test set, and the MSA diacritizer produced significantly lower results with a WER of 8.5\% (see Table \ref{res:coreWordComparison}).  This highlights the large difference between MSA and CA and the need for systems that are specifically tuned for both.

We randomly selected and analyzed 500 errors (5.2\% of the errors).  The errors types along with examples of each are shown in Table \ref{table:CWErrorTypesCA}. The two most common errors involve the system producing completely correct diacritized forms (38.8\%) or correct forms that don't match the context (31.4\%).  The relatively higher percentage of completely incorrect guesses, compared to MSA, may point to the higher lexical diversity of classical Arabic. As for MSA, we suspect that adding additional POS information and employing a word bigram to constrain the PRIOR feature may help reduce selection errors. Another prominent error is related to the diacritics that appear on attached suffixes, particularly pronouns, which depend on the choice of case ending (13.2\%).  Errors due to named entities are slightly fewer than those seen for MSA (8.8\%).  A noticeable number of mismatches between the guess and the reference are due to partial diacritization of the reference (4.4\%). 
We plan to conduct an extra round of checks on the test set.  

\begin{table*}[hbt!]
\begin{center}
\begin{tabular}{p{2.2cm}|r|r|p{3.8cm}|p{3.8cm}}
Error & Freq. & \% & Explanation & Examples \\ \hline
Invalid diacritized form & 195 & 38.8 & invalid form & ``$>$aqosaAm'' (\<أِقْسَام> -- portions) vs. ``$>$aqasaAm'' (\<أَقَسَام>)\\ \hline
Wrong selection & 157 & 31.4 & Homographs with different diacritized forms & ``raAfoE'' (\<رَقْع> -- lifting) vs. ``rafaE'' (\<رَفَع> -- he lifted) \\ \hline
Affix diacritization error & 66 & 13.2 & Some affixes are erroneously diacritized & ``baladhu'' (\<بَلَدهُ> -- his country, where country is subject of verb) vs. ``baladhi'' (\<بَلَدهِ> -- his country, where country is subject or object of preposition) \\ \hline
Named entities & 44 & 8.8 & Named entities & ``Alr\textasciitilde{}ayob'' (\<الرَّيْب> -- Arrayb) vs. ``Alr\textasciitilde{}iyab'' (\<الرِّيَب>)) \\ \hline
Problems with reference & 22 & 4.4 & Some words in the reference were partially diacritized & ``nuEoTaY'' (\<نُعْطَى> -- we are given) vs. ``nETY'' (\<نعطى>)) \\ \hline
Guess has no diacritics & 9 & 1.8 & system did not produce any diacritics & ``mhnd'' (\<مهند> -- sword) vs. ``muhan\textasciitilde{}ad'' (\<مُهَنَّد>)) \\ \hline
Different valid forms & 7 & 1.4 & Some words have multiple valid diacritized forms & ``maA\}op''  (\<مَائْة> -- hundred) and ``miA\}op'' (\<مِائَة>) \\ \hline
Misspelled word & 1 & 0.2 & & ``lbAlmsjd'' (\<لبالمسجد>) vs. ``lbAlmsjd'' (\<بالمسجد> -- in the mosque)) \\ \hline
\end{tabular}
\caption{Error analysis: Core word error types for CA}
\label{table:CWErrorTypesCA}
\end{center}
\end{table*}

\begin{table}[hbt!]
\begin{center}
\begin{tabular}{l|c|c}
 & \multicolumn{2}{c}{Error Rate} \\
System & WER & DER \\ \hline
\multicolumn{3}{c}{MSA} \\ \hline
\textbf{Our system} & \textbf{2.9} & \textbf{0.9} \\
\citep{rashwan2015deep} & 3.0 & 1.0 \\
Farasa & 3.3 & 1.1 \\
Microsoft ATKS	&	5.7 & 2.0	\\
MADAMIRA & 6.7 & 1.9 \\ 
\citep{belinkov2015arabic} & 14.9 & 3.9 \\ \hline
\multicolumn{3}{c}{CA} \\ \hline
Our system & 2.2 & 0.9 \\
Our best MSA system on CA & 8.5 & 3.7 \\
\end{tabular}
\caption{Comparing our system to state-of-the-art systems -- Core word diacritics}
\label{res:coreWordComparison}
\end{center}
\end{table}

\subsection{Case Ending}
\subsubsection{Experimental Setup}
We conducted multiple experiments to determine the relative effect of the different features as follows: 
\begin{itemize}
\item \textbf{word}: This is our baseline setup, which uses word surface forms only.
\item \textbf{word-surface}: This setup uses the word surface forms, stems, prefixes, and suffixes (including noun suffixes).  This simulates the case when no POS tagging information is available.
\item  \textbf{word-POS}: This includes the word surface form and POS information, including gender and number of stems, prefixes, and suffixes.
\item \textbf{word-morph}: This includes words and their stem templates to capture morphological patterns.
\item \textbf{word-surface-POS-morph}: This setup uses all the features (surface, POS, and morphological). 
\item \textbf{all-misc}: This uses all features plus word and stem leading and trailing character unigrams and bigrams in addition to \textit{sukun} words and named entities.
\end{itemize}

For testing MSA, we used the aforementioned WikiNews dataset.  Again, we compared our results to five other systems, namely Farasa~\citep{darwish2017arabic},
MADAMIRA~\citep{pasha2014madamira}, 
RDI (Rashwan et al., 2015), MIT (Belinkow and Glass, 2015), and Microsoft ATKS \citep{microsoft2013diac}. 
For CA testing, we used the 5,000 sentences that we set aside.  Again, we compared to our best MSA system. 

\subsubsection{Results and Error Analysis}
Table \ref{table:results} lists the results of our setups 
compared to other systems.  

\paragraph{\textbf{MSA Results:}} As the results show, our baseline DNN system outperforms all state-of-the-art systems.  Further, adding more features yielded better results overall.  Surface-level features resulted in the most gain, followed by POS tags, and lastly 
stem templates.  Further, adding head and tail characters along with a list of \textit{sukun} words and named entities led to further improvement. Our proposed feature-rich system has a CEER that is approximately 61\% lower than any of the state-of-the-art systems.

\begin{figure}[]
\begin{center}
\includegraphics[width=0.9\linewidth]{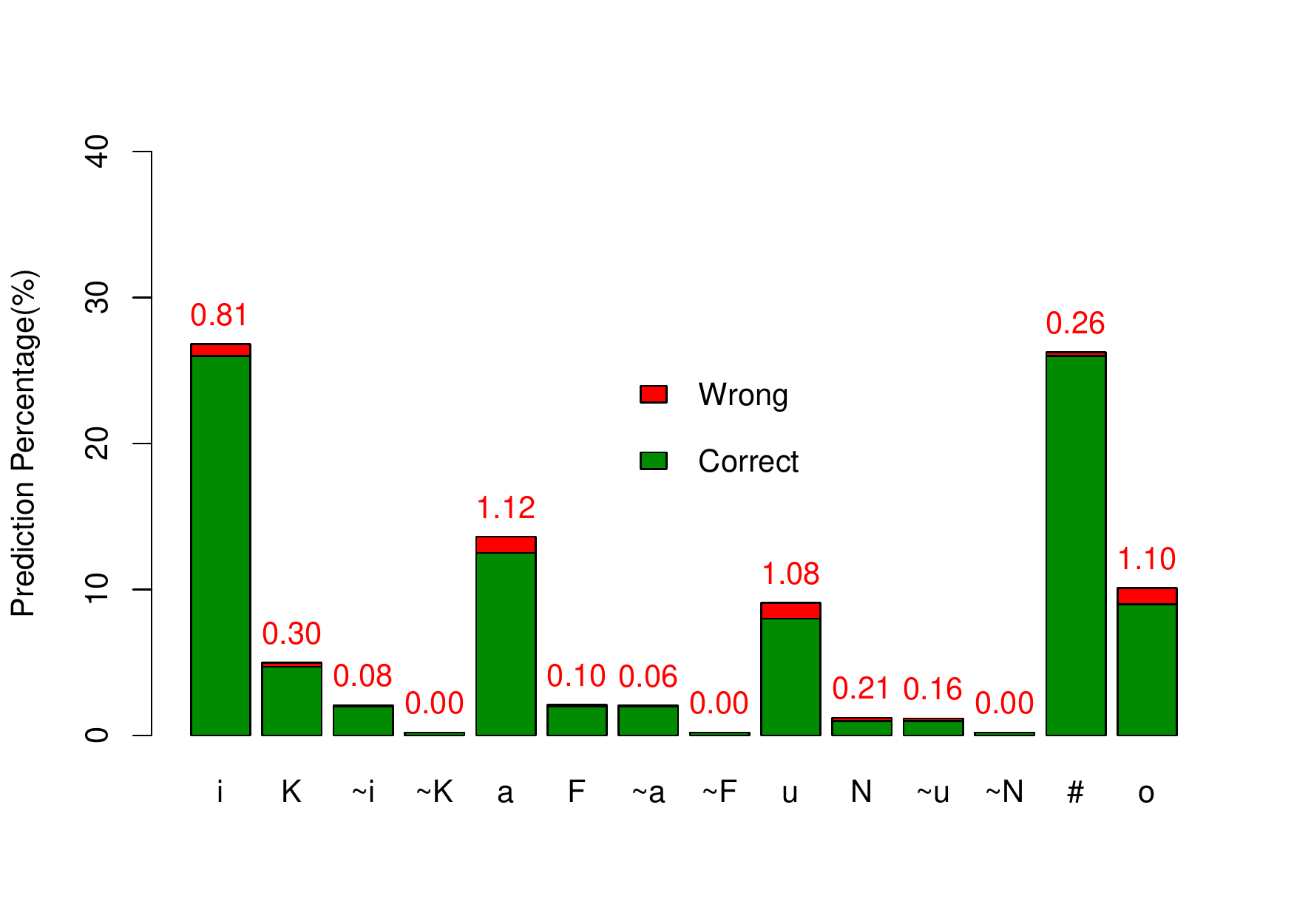}
\caption{Case endings distribution and prediction accuracy for MSA}
\label{fig:pred-accu-MSA}
\end{center}
\end{figure}

Figure \ref{fig:pred-accu-MSA} shows CE distribution and prediction accuracy. For the four basic markers \textit{kasra}, \textit{fatHa}, \textit{damma} and \textit{sukun}, which appear 27\%, 14\%, 9\% and 10\% respectively, the system has CEER of $\sim$1\% for each. Detecting the virtual CE mark is a fairly easy task. All other CE markers represent 13\% with almost negligible errors. 

Table \ref{tab:msaCaseErrors} lists a thorough breakdown of all errors accounting for at 1\% of the errors along with the most  common reasons of the errors and examples illustrating these reasons.  For example, the most common error type involves guessing a fatHa (a) instead of damma (u) or vice versa (19.3\%).  The most common reasons for this error type, based on inspecting the errors, were due to: POS errors (ex. a word is tagged as a verb instead of a noun); and a noun is treated as a subject instead of an object or vice versa.  The table details the rest of the error types.  
Overall, some of the errors are potentially fixable using better POS tagging, improved detection of non-Arabized foreign names, and detection of indeclinability.  However, some errors are more difficult and require greater understanding of semantics such as improper attachment, incorrect idafa, and confusion between subject and object.  Perhaps, such semantic errors can be resolved using parsing.


\begin{table}[hbt!]
\begin{center}
\begin{tabular}{l|r}
Setup & CEER\% \\ \hline
\multicolumn{2}{c}{MSA} \\ \hline
word (baseline) &   9.1 \\
word-surface 	&	5.7 \\
word-POS		&	7.0 \\
word-morph		&	7.6 \\
word-surface-POS-morph				&	5.2 \\
\textbf{all-misc} 		& \textbf{3.7} \\ \hline
Microsoft ATKS	& 9.5	\\
Farasa			& 10.4	\\
RDI \citep{rashwan2015deep} & 14.0	\\
MIT \citep{belinkov2015arabic} & 15.3	\\
MADAMIRA \citep{pasha2014madamira} & 15.9	\\ \hline
\multicolumn{2}{c}{CA} \\ \hline
word (baseline) &   4.0 \\
word-surface 	&	3.3 \\
word-POS		&	3.1 \\
word-morph		&	3.7 \\
word-surface-POS-morph				&	2.9 \\
\textbf{all-misc} 		& \textbf{2.5} \\ \hline
Our best MSA system on CA & 8.9 \\
\end{tabular}
\caption{MSA Results and comparison to other systems}
\label{table:results}
\end{center}
\end{table}

\paragraph{\textbf{CA Results:}}  The results show that the POS tagging features led to the most improvements followed by the surface features.  Combining all features led to the best results with WER of 2.5\%.  As we saw for CW diacritics, using our best MSA system to diacritize CA led to significantly lower results with CEER of 8.9\%.

\begin{figure}[]
\begin{center}
\includegraphics[width=0.9\linewidth]{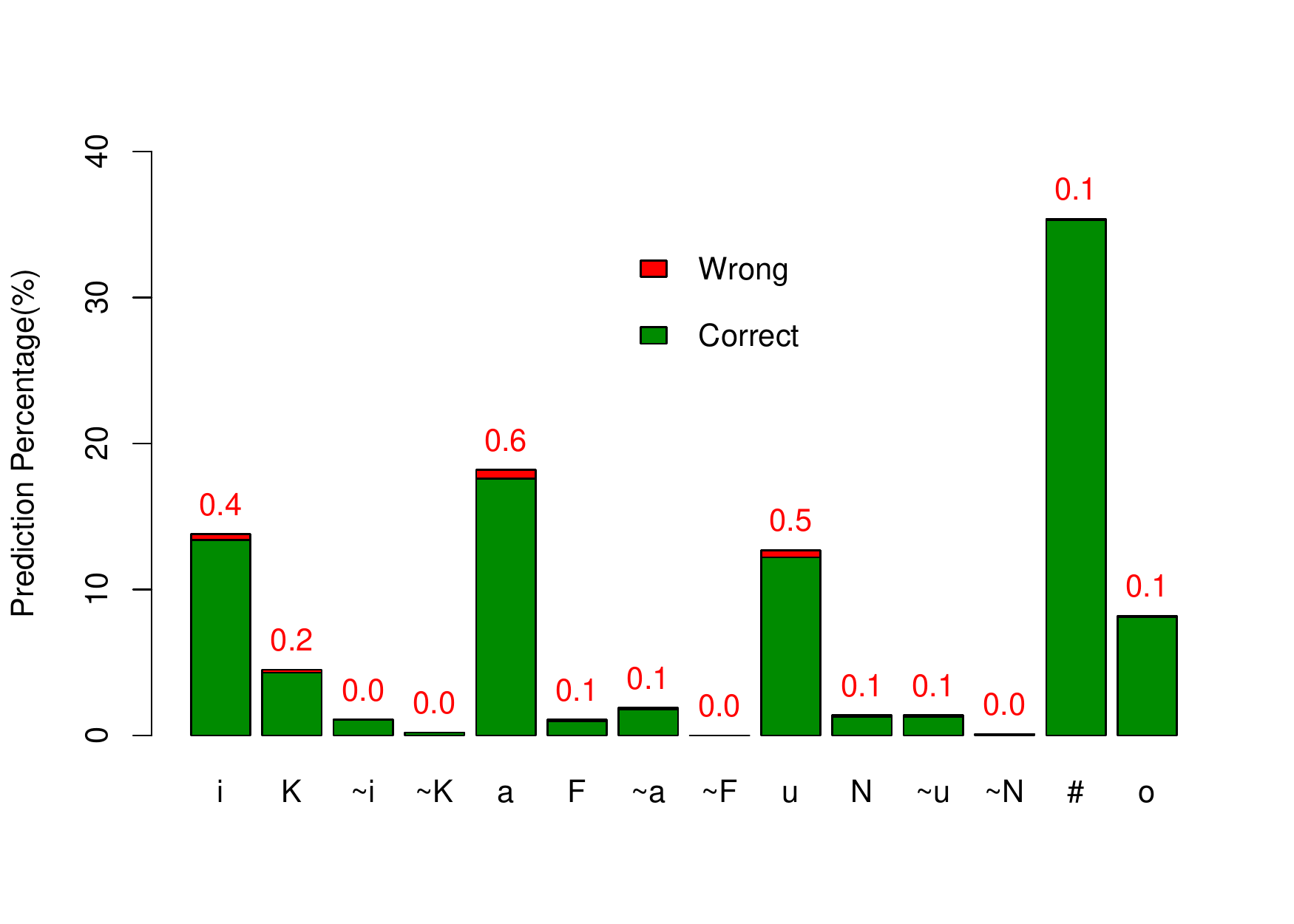}
\caption{Case endings distribution and prediction accuracy for CA}
\label{fig:pred-accu-CA}
\end{center}
\end{figure}

Figure \ref{fig:pred-accu-CA} shows CE distribution and prediction accuracy. For the four basic markers \textit{fatHa}, \textit{kasra}, \textit{damma} and \textit{sukun}, which appear 18\%, 14\%, 13\% and 8\% respectively, the system has CEER $\sim$0.5\% for each. Again, detecting the virtual CE mark was a fairly easy task. All other CE markers representing 20\% have negligible  errors.


Table \ref{tab:caCaseErrors} lists all the error types, which account for at least 1\% of the errors, along with their most common causes and explanatory examples.  The error types are similar to those observed for MSA.  Some errors are more syntactic and morphological in nature and can be addressed using better POS tagging and identification of indeclinability, particularly as they relate to named entities and nouns with feminine markers.  Other errors such as incorrect attachment, incorrect idafa, false subject, and confusion between subject and object can perhaps benefit from the use of parsing.  As with the core-word errors for CA, the reference has some errors (ex. \{a,i,o\} $\Rightarrow$ \#), and extra rounds of reviews of the reference are in order. 





\subsection{Full Diacritization Results}
Table \ref{table:resultsCombined} compares the full word diacritization (CW+CE) of our best setup to other systems in the literature.  As the results show for MSA, our overall diacritization WER is 6.0\% while the 
state of the art
system has a WER of 12.2\%.  As for CA, our best system produced an error rate of 4.3\%, which is significantly better than using our best MSA system to diacritize CA.

\begin{table}[hbt!]
\begin{center}
\begin{tabular}{l|r}
Setup & WER\% \\ \hline
\multicolumn{2}{c}{MSA} \\ \hline
\textbf{Our System} 		&   \textbf{6.0} \\ \hline
Microsoft ATKS	&	12.2 \\
Farasa			&	12.8 \\
RDI \citep{rashwan2015deep} &	16.0	\\
MADAMIRA \citep{pasha2014madamira} &	19.0	\\
MIT \citep{belinkov2015arabic} &	30.5	\\ \hline
\multicolumn{2}{c}{CA} \\ \hline
Our system & \textbf{4.3} \\
Our best MSA system on CA & 14.7 \\
\end{tabular}
\caption{Comparison to other systems for full diacritization}
\label{table:resultsCombined}
\end{center}
\end{table}

\section{Conclusion and Future Work}
In this paper, we presented a feature-rich DNN approach for MSA CW and CE recovery that produces a word level error for MSA of 6.0\%, which is more than 50\% lower than state-of-the-art systems (6.0\% compared to 12.2\%) and word error rate of 4.3\% for CA.  Specifically, we used biLSTM-based model with a variety of surface, morphological, and syntactic features. Reliable NLP tools may be required to generate some of these features, and such tools may not be readily available for other language varieties, such as dialectal Arabic.  However, we showed the efficacy of different varieties of features, such as surface level-features, and they can help improve diacritization individually. 
Further, though some errors may be overcome using improved NLP tools (ex. better POS tagging), semantic errors, such incorrect attchment, are more difficult to fix.  Perhaps, using dependency parsing may help overcome some semantic errors.  As for feature engineering, the broad categories of features, such as surface, syntactic, and morphological features, may likely carry-over to other languages, language-specific feature engineering may be require to handle the specificity of each language.  Lastly, since multiple diacritization conventions may exist, as in the case of Arabic, adopting one convention consistently is important for training a good system and for properly testing it.  Though we have mostly achieved this for MSA, the CA dataset requires more checks to insure greater consistency.  

For future work, we want to explore the effectiveness of augmenting our CW model with POS tagging information and a bigram language model.  Further, we plan to create a multi reference diacritization test set to handle words that have multiple valid diacritized forms. For CE, we want to examine the effectiveness of the proposed features for Arabic parsing. We plan to explore: character-level convolutional neural networks that may capture sub-word morphological features; pre-trained embeddings; and attention mechanisms to focus on salient features. We also plan to explore joint modeling for both core word and case ending diacritics.





\bibliographystyle{acl_natbib}
\bibliography{nlebib.bib}

\begin{thebibliography}{}

\bibitem[\protect\citeauthoryear{Abadi \bgroup et al\mbox.\egroup
  }{2015}]{tensorflow2015-whitepaper}
Abadi, M.; Agarwal, A.; Barham, P.; Brevdo, E.; Chen, Z.; Citro, C.; Corrado,
  G.~S.; Davis, A.; Dean, J.; Devin, M.; Ghemawat, S.; Goodfellow, I.; Harp,
  A.; Irving, G.; Isard, M.; Jia, Y.; Jozefowicz, R.; Kaiser, L.; Kudlur, M.;
  Levenberg, J.; Man\'{e}, D.; Monga, R.; Moore, S.; Murray, D.; Olah, C.;
  Schuster, M.; Shlens, J.; Steiner, B.; Sutskever, I.; Talwar, K.; Tucker, P.;
  Vanhoucke, V.; Vasudevan, V.; Vi\'{e}gas, F.; Vinyals, O.; Warden, P.;
  Wattenberg, M.; Wicke, M.; Yu, Y.; and Zheng, X.
\newblock 2015.
\newblock {TensorFlow}: Large-scale machine learning on heterogeneous systems.
\newblock Software available from tensorflow.org.

\bibitem[\protect\citeauthoryear{Abandah \bgroup et al\mbox.\egroup
  }{2015}]{abandah2015automatic}
Abandah, G.~A.; Graves, A.; Al-Shagoor, B.; Arabiyat, A.; Jamour, F.; and
  Al-Taee, M.
\newblock 2015.
\newblock Automatic diacritization of arabic text using recurrent neural
  networks.
\newblock {\em International Journal on Document Analysis and Recognition
  (IJDAR)} 18(2):183--197.

\bibitem[\protect\citeauthoryear{Ameur, Moulahoum, and
  Guessoum}{2015}]{ameur2015restoration}
Ameur, M. S.~H.; Moulahoum, Y.; and Guessoum, A.
\newblock 2015.
\newblock Restoration of arabic diacritics using a multilevel statistical
  model.
\newblock In {\em IFIP International Conference on Computer Science and its
  Applications\_x000D\_},  181--192.
\newblock Springer.

\bibitem[\protect\citeauthoryear{Attia}{2008}]{Attia:08}
Attia, M.
\newblock 2008.
\newblock {\em Handling Arabic morphological and syntactic ambiguity within the
  LFG framework with a view to machine translation}.
\newblock The University of Manchester, UK: Ph.D. Thesis. School of Languages,
  Linguistics and Cultures.

\bibitem[\protect\citeauthoryear{Azmi and Almajed}{2015}]{azmi2015survey}
Azmi, A.~M., and Almajed, R.~S.
\newblock 2015.
\newblock A survey of automatic arabic diacritization techniques.
\newblock {\em Natural Language Engineering} 21(03):477--495.

\bibitem[\protect\citeauthoryear{Bebah \bgroup et al\mbox.\egroup
  }{2014}]{bebah2014hybrid}
Bebah, M.; Amine, C.; Azzeddine, M.; and Abdelhak, L.
\newblock 2014.
\newblock Hybrid approaches for automatic vowelization of arabic texts.
\newblock {\em arXiv preprint arXiv:1410.2646}.

\bibitem[\protect\citeauthoryear{Belinkov and Glass}{2015}]{belinkov2015arabic}
Belinkov, Y., and Glass, J.
\newblock 2015.
\newblock Arabic diacritization with recurrent neural networks.
\newblock In {\em Proceedings of the 2015 Conference on Empirical Methods in
  Natural Language Processing},  2281--2285.

\bibitem[\protect\citeauthoryear{Buckwalter}{2004}]{buckwalter2004buckwalter}
Buckwalter, T.
\newblock 2004.
\newblock Buckwalter arabic morphological analyzer version 2.0.
\newblock {\em LDC catalog number LDC2004L02, ISBN 1-58563-324-0.}

\bibitem[\protect\citeauthoryear{Chollet and others}{2015}]{chollet2015keras}
Chollet, F., et~al.
\newblock 2015.
\newblock Keras.
\newblock \url{https://keras.io}.

\bibitem[\protect\citeauthoryear{Darwish and Gao}{2014}]{darwish2014simple}
Darwish, K., and Gao, W.
\newblock 2014.
\newblock Simple effective microblog named entity recognition: Arabic as an
  example.
\newblock In {\em LREC},  2513--2517.

\bibitem[\protect\citeauthoryear{Darwish and Mubarak}{2016}]{DARWISH2016farasa}
Darwish, K., and Mubarak, H.
\newblock 2016.
\newblock Farasa: A new fast and accurate arabic word segmenter.
\newblock In {\em Proceedings of the Tenth International Conference on Language
  Resources and Evaluation (LREC 2016)}.
\newblock Paris, France: European Language Resources Association (ELRA).

\bibitem[\protect\citeauthoryear{Darwish, Mubarak, and
  Abdelali}{2017}]{darwish2017arabic}
Darwish, K.; Mubarak, H.; and Abdelali, A.
\newblock 2017.
\newblock Arabic diacritization: Stats, rules, and hacks.
\newblock In {\em Proceedings of the Third Arabic Natural Language Processing
  Workshop},  9--17.

\bibitem[\protect\citeauthoryear{Darwish}{2013}]{darwish2013named}
Darwish, K.
\newblock 2013.
\newblock Named entity recognition using cross-lingual resources: Arabic as an
  example.
\newblock In {\em Proceedings of the 51st Annual Meeting of the Association for
  Computational Linguistics (Volume 1: Long Papers)}, volume~1,  1558--1567.

\bibitem[\protect\citeauthoryear{El-Sadany and Hashish}{1989}]{el1989arabic}
El-Sadany, T.~A., and Hashish, M.~A.
\newblock 1989.
\newblock An arabic morphological system.
\newblock {\em IBM Systems Journal} 28(4):600--612.

\bibitem[\protect\citeauthoryear{Elshafei, Al-Muhtaseb, and
  Alghamdi}{2006}]{elshafei2006statistical}
Elshafei, M.; Al-Muhtaseb, H.; and Alghamdi, M.
\newblock 2006.
\newblock Statistical methods for automatic diacritization of arabic text.
\newblock In {\em The Saudi 18th National Computer Conference. Riyadh},
  volume~18,  301--306.

\bibitem[\protect\citeauthoryear{Gal}{2002}]{gal2002hmm}
Gal, Y.
\newblock 2002.
\newblock An hmm approach to vowel restoration in arabic and hebrew.
\newblock In {\em Proceedings of the ACL-02 workshop on Computational
  approaches to Semitic languages},  1--7.
\newblock Association for Computational Linguistics.

\bibitem[\protect\citeauthoryear{Habash and Rambow}{2007}]{habash2007arabic}
Habash, N., and Rambow, O.
\newblock 2007.
\newblock Arabic diacritization through full morphological tagging.
\newblock In {\em Human Language Technologies 2007: The Conference of the North
  American Chapter of the Association for Computational Linguistics; Companion
  Volume, Short Papers},  53--56.
\newblock Association for Computational Linguistics.

\bibitem[\protect\citeauthoryear{Harrat \bgroup et al\mbox.\egroup
  }{2013}]{harrat:hal-00925815}
Harrat, S.; Abbas, M.; Meftouh, K.; and Smaili, K.
\newblock 2013.
\newblock {Diacritics Restoration for Arabic Dialects}.
\newblock In {\em {INTERSPEECH 2013 - 14th Annual Conference of the
  International Speech Communication Association}}.
\newblock Lyon, France: {ISCA}.

\bibitem[\protect\citeauthoryear{Hinton \bgroup et al\mbox.\egroup
  }{2012}]{hinton2012improving}
Hinton, G.~E.; Srivastava, N.; Krizhevsky, A.; Sutskever, I.; and
  Salakhutdinov, R.~R.
\newblock 2012.
\newblock Improving neural networks by preventing co-adaptation of feature
  detectors.
\newblock {\em arXiv preprint arXiv:1207.0580}.

\bibitem[\protect\citeauthoryear{Maamouri \bgroup et al\mbox.\egroup
  }{2004}]{maamouri2004atb3}
Maamouri, M.; Bies, A.; Buckwalter, T.; and Mekki, W.
\newblock 2004.
\newblock The penn arabic treebank: building a large-scale annotated arabic
  corpus.
\newblock In {\em NEMLAR Conference on Arabic Language Resources and Tools},
  102–--109.

\bibitem[\protect\citeauthoryear{Marton, Habash, and
  Rambow}{2010}]{marton2010improving}
Marton, Y.; Habash, N.; and Rambow, O.
\newblock 2010.
\newblock Improving arabic dependency parsing with lexical and inflectional
  morphological features.
\newblock In {\em Proceedings of the NAACL HLT 2010 First Workshop on
  Statistical Parsing of Morphologically-Rich Languages},  13--21.
\newblock Association for Computational Linguistics.

\bibitem[\protect\citeauthoryear{Mohamed Ould Abdallahi~Ould \bgroup et
  al\mbox.\egroup }{2011}]{bebah2011alkhalil}
Mohamed Ould Abdallahi~Ould, B.; Meziane, A.; Mazroui, A.; and Lakhouaja, A.
\newblock 2011.
\newblock Alkhalil morphosys.
\newblock In {\em 7th International Computing Conference in Arabic},  66--73.

\bibitem[\protect\citeauthoryear{Nelken and Shieber}{2005}]{nelken2005arabic}
Nelken, R., and Shieber, S.~M.
\newblock 2005.
\newblock Arabic diacritization using weighted finite-state transducers.
\newblock In {\em Proceedings of the ACL Workshop on Computational Approaches
  to Semitic Languages},  79--86.
\newblock Association for Computational Linguistics.

\bibitem[\protect\citeauthoryear{Osama~Hamed}{2017}]{HamedZesch:2017}
Osama~Hamed, T.~Z.
\newblock 2017.
\newblock {A Survey and Comparative Study of Arabic Diacritization Tools}.
\newblock {\em JLCL} 32(1):27--47.

\bibitem[\protect\citeauthoryear{Pasha \bgroup et al\mbox.\egroup
  }{2014}]{pasha2014madamira}
Pasha, A.; Al-Badrashiny, M.; Diab, M.; El~Kholy, A.; Eskander, R.; Habash, N.;
  Pooleery, M.; Rambow, O.; and Roth, R.~M.
\newblock 2014.
\newblock Madamira: A fast, comprehensive tool for morphological analysis and
  disambiguation of arabic.
\newblock In {\em LREC-2014}.

\bibitem[\protect\citeauthoryear{Rashwan \bgroup et al\mbox.\egroup
  }{2015}]{rashwan2015deep}
Rashwan, M.; Al~Sallab, A.; Raafat, M.; and Rafea, A.
\newblock 2015.
\newblock Deep learning framework with confused sub-set resolution architecture
  for automatic arabic diacritization.
\newblock In {\em IEEE Transactions on Audio, Speech, and Language Processing},
   505--516.

\bibitem[\protect\citeauthoryear{Roth \bgroup et al\mbox.\egroup
  }{2008}]{roth2008arabic}
Roth, R.; Rambow, O.; Habash, N.; Diab, M.; and Rudin, C.
\newblock 2008.
\newblock Arabic morphological tagging, diacritization, and lemmatization using
  lexeme models and feature ranking.
\newblock In {\em Proceedings of the 46th Annual Meeting of the Association for
  Computational Linguistics on Human Language Technologies: Short Papers},
  117--120.
\newblock Association for Computational Linguistics.

\bibitem[\protect\citeauthoryear{Said \bgroup et al\mbox.\egroup
  }{2013}]{microsoft2013diac}
Said, A.; El-Sharqwi, M.; Chalabi, A.; and Kamal, E.
\newblock 2013.
\newblock A hybrid approach for arabic diacritization.
\newblock In M{\'e}tais, E.; Meziane, F.; Saraee, M.; Sugumaran, V.; and
  Vadera, S., eds., {\em Natural Language Processing and Information Systems},
  53--64.
\newblock Berlin, Heidelberg: Springer Berlin Heidelberg.

\bibitem[\protect\citeauthoryear{Schuster and
  Paliwal}{1997}]{schuster1997bilstm}
Schuster, M., and Paliwal, K.~K.
\newblock 1997.
\newblock Bidirectional recurrent neural networks.
\newblock {\em IEEE Transactions on Signal Processing} 45(11):2673--2681.

\bibitem[\protect\citeauthoryear{Vergyri and
  Kirchhoff}{2004}]{vergyri2004automatic}
Vergyri, D., and Kirchhoff, K.
\newblock 2004.
\newblock Automatic diacritization of arabic for acoustic modeling in speech
  recognition.
\newblock In {\em Proceedings of the workshop on computational approaches to
  Arabic script-based languages, COLING'04},  66--73.
\newblock Geneva, Switzerland: Association for Computational Linguistics.

\bibitem[\protect\citeauthoryear{Zitouni, Sorensen, and
  Sarikaya}{2006}]{zitouni2006maximum}
Zitouni, I.; Sorensen, J.~S.; and Sarikaya, R.
\newblock 2006.
\newblock Maximum entropy based restoration of arabic diacritics.
\newblock In {\em Proceedings of the 21st International Conference on
  Computational Linguistics and the 44th annual meeting of the Association for
  Computational Linguistics},  577--584.
\newblock Association for Computational Linguistics.

\end{thebibliography}


\begin{thebibliography}{43}
\expandafter\ifx\csname natexlab\endcsname\relax\def\natexlab#1{#1}\fi

\bibitem[{Abadi et~al.(2015)Abadi, Agarwal, Barham, Brevdo, Chen, Citro,
  Corrado, Davis, Dean, Devin, Ghemawat, Goodfellow, Harp, Irving, Isard, Jia,
  Jozefowicz, Kaiser, Kudlur, Levenberg, Man\'{e}, Monga, Moore, Murray, Olah,
  Schuster, Shlens, Steiner, Sutskever, Talwar, Tucker, Vanhoucke, Vasudevan,
  Vi\'{e}gas, Vinyals, Warden, Wattenberg, Wicke, Yu, and
  Zheng}]{tensorflow2015-whitepaper}
Mart\'{\i}n Abadi, Ashish Agarwal, Paul Barham, Eugene Brevdo, Zhifeng Chen,
  Craig Citro, Greg~S. Corrado, Andy Davis, Jeffrey Dean, Matthieu Devin,
  Sanjay Ghemawat, Ian Goodfellow, Andrew Harp, Geoffrey Irving, Michael Isard,
  Yangqing Jia, Rafal Jozefowicz, Lukasz Kaiser, Manjunath Kudlur, Josh
  Levenberg, Dandelion Man\'{e}, Rajat Monga, Sherry Moore, Derek Murray, Chris
  Olah, Mike Schuster, Jonathon Shlens, Benoit Steiner, Ilya Sutskever, Kunal
  Talwar, Paul Tucker, Vincent Vanhoucke, Vijay Vasudevan, Fernanda Vi\'{e}gas,
  Oriol Vinyals, Pete Warden, Martin Wattenberg, Martin Wicke, Yuan Yu, and
  Xiaoqiang Zheng. 2015.
\newblock {TensorFlow}: Large-scale machine learning on heterogeneous systems.
\newblock Software available from tensorflow.org.

\bibitem[{Abandah et~al.(2015)Abandah, Graves, Al-Shagoor, Arabiyat, Jamour,
  and Al-Taee}]{abandah2015automatic}
Gheith~A Abandah, Alex Graves, Balkees Al-Shagoor, Alaa Arabiyat, Fuad Jamour,
  and Majid Al-Taee. 2015.
\newblock Automatic diacritization of arabic text using recurrent neural
  networks.
\newblock \emph{International Journal on Document Analysis and Recognition
  (IJDAR)}, 18(2):183--197.

\bibitem[{Ameur et~al.(2015)Ameur, Moulahoum, and
  Guessoum}]{ameur2015restoration}
Mohamed Seghir~Hadj Ameur, Youcef Moulahoum, and Ahmed Guessoum. 2015.
\newblock Restoration of arabic diacritics using a multilevel statistical
  model.
\newblock In \emph{IFIP International Conference on Computer Science and its
  Applications\_x000D\_}, pages 181--192. Springer.

\bibitem[{Attia(2008)}]{Attia:08}
Mohammed Attia. 2008.
\newblock \emph{Handling Arabic morphological and syntactic ambiguity within
  the LFG framework with a view to machine translation}.
\newblock Ph.D. Thesis. School of Languages, Linguistics and Cultures, The
  University of Manchester, UK.

\bibitem[{Azmi and Almajed(2015)}]{azmi2015survey}
Aqil~M Azmi and Reham~S Almajed. 2015.
\newblock A survey of automatic arabic diacritization techniques.
\newblock \emph{Natural Language Engineering}, 21(03):477--495.

\bibitem[{Bebah et~al.(2014)Bebah, Amine, Azzeddine, and
  Abdelhak}]{bebah2014hybrid}
Mohamed Bebah, Chennoufi Amine, Mazroui Azzeddine, and Lakhouaja Abdelhak.
  2014.
\newblock Hybrid approaches for automatic vowelization of arabic texts.
\newblock \emph{arXiv preprint arXiv:1410.2646}.

\bibitem[{Belinkov and Glass(2015)}]{belinkov2015arabic}
Yonatan Belinkov and James Glass. 2015.
\newblock Arabic diacritization with recurrent neural networks.
\newblock In \emph{Proceedings of the 2015 Conference on Empirical Methods in
  Natural Language Processing}, pages 2281--2285, Lisbon, Portugal.

\bibitem[{Buckwalter(2002)}]{buckwalter2002buckwalter}
Tim Buckwalter. 2002.
\newblock Buckwalter $\{$Arabic$\}$ morphological analyzer version 1.0.
\newblock \emph{LDC catalog number LDC2002L49, ISBN 1-58563-257-0.}

\bibitem[{Buckwalter(2004)}]{buckwalter2004buckwalter}
Tim Buckwalter. 2004.
\newblock Buckwalter arabic morphological analyzer version 2.0.
\newblock \emph{LDC catalog number LDC2004L02, ISBN 1-58563-324-0.}

\bibitem[{Chollet et~al.(2015)}]{chollet2015keras}
Fran\c{c}ois Chollet et~al. 2015.
\newblock Keras.
\newblock \url{https://keras.io}.

\bibitem[{Darwish(2013)}]{darwish2013named}
Kareem Darwish. 2013.
\newblock Named entity recognition using cross-lingual resources: Arabic as an
  example.
\newblock In \emph{Proceedings of the 51st Annual Meeting of the Association
  for Computational Linguistics (Volume 1: Long Papers)}, volume~1, pages
  1558--1567.

\bibitem[{Darwish et~al.(2018)Darwish, Abdelali, Mubarak, Samih, and
  Attia}]{darwish2018diacritization}
Kareem Darwish, Ahmed Abdelali, Hamdy Mubarak, Younes Samih, and Mohammed
  Attia. 2018.
\newblock Diacritization of moroccan and tunisian arabic dialects: A crf
  approach.
\newblock In \emph{OSACT 3: The 3rd Workshop on Open-Source Arabic Corpora and
  Processing Tools}, page~62.

\bibitem[{Darwish and Gao(2014)}]{darwish2014simple}
Kareem Darwish and Wei Gao. 2014.
\newblock Simple effective microblog named entity recognition: Arabic as an
  example.
\newblock In \emph{LREC}, pages 2513--2517.

\bibitem[{Darwish and Mubarak(2016)}]{DARWISH2016farasa}
Kareem Darwish and Hamdy Mubarak. 2016.
\newblock Farasa: A new fast and accurate arabic word segmenter.
\newblock In \emph{Proceedings of the Tenth International Conference on
  Language Resources and Evaluation (LREC 2016)}, Paris, France. European
  Language Resources Association (ELRA).

\bibitem[{Darwish et~al.(2017)Darwish, Mubarak, and
  Abdelali}]{darwish2017arabic}
Kareem Darwish, Hamdy Mubarak, and Ahmed Abdelali. 2017.
\newblock Arabic diacritization: Stats, rules, and hacks.
\newblock In \emph{Proceedings of the Third Arabic Natural Language Processing
  Workshop}, pages 9--17.

\bibitem[{De~Pauw et~al.(2007)De~Pauw, Wagacha, and
  De~Schryver}]{de2007automatic}
Guy De~Pauw, Peter~W Wagacha, and Gilles-Maurice De~Schryver. 2007.
\newblock Automatic diacritic restoration for resource-scarce languages.
\newblock In \emph{International Conference on Text, Speech and Dialogue},
  pages 170--179. Springer.

\bibitem[{El-Sadany and Hashish(1989)}]{el1989arabic}
Tarek~A. El-Sadany and Mohamed~A. Hashish. 1989.
\newblock An arabic morphological system.
\newblock \emph{IBM Systems Journal}, 28(4):600--612.

\bibitem[{Elshafei et~al.(2006)Elshafei, Al-Muhtaseb, and
  Alghamdi}]{elshafei2006statistical}
Moustafa Elshafei, Husni Al-Muhtaseb, and Mansour Alghamdi. 2006.
\newblock Statistical methods for automatic diacritization of arabic text.
\newblock In \emph{The Saudi 18th National Computer Conference. Riyadh},
  volume~18, pages 301--306.

\bibitem[{Gal(2002)}]{gal2002hmm}
Ya'akov Gal. 2002.
\newblock An hmm approach to vowel restoration in arabic and hebrew.
\newblock In \emph{Proceedings of the ACL-02 workshop on Computational
  approaches to Semitic languages}, pages 1--7. Association for Computational
  Linguistics.

\bibitem[{Habash and Rambow(2007)}]{habash2007arabic}
Nizar Habash and Owen Rambow. 2007.
\newblock Arabic diacritization through full morphological tagging.
\newblock In \emph{Human Language Technologies 2007: The Conference of the
  North American Chapter of the Association for Computational Linguistics;
  Companion Volume, Short Papers}, pages 53--56. Association for Computational
  Linguistics.

\bibitem[{Harrat et~al.(2013)Harrat, Abbas, Meftouh, and
  Smaili}]{harrat:hal-00925815}
Salima Harrat, Mourad Abbas, Karima Meftouh, and Kamel Smaili. 2013.
\newblock {Diacritics Restoration for Arabic Dialects}.
\newblock In \emph{{INTERSPEECH 2013 - 14th Annual Conference of the
  International Speech Communication Association}}, Lyon, France. {ISCA}.

\bibitem[{{Hifny}(2018)}]{Hifny2018Hybrid}
Y.~{Hifny}. 2018.
\newblock Hybrid lstm/maxent networks for arabic syntactic diacritics
  restoration.
\newblock \emph{IEEE Signal Processing Letters}, 25(10):1515--1519.

\bibitem[{Hinton et~al.(2012)Hinton, Srivastava, Krizhevsky, Sutskever, and
  Salakhutdinov}]{hinton2012improving}
Geoffrey~E Hinton, Nitish Srivastava, Alex Krizhevsky, Ilya Sutskever, and
  Ruslan~R Salakhutdinov. 2012.
\newblock Improving neural networks by preventing co-adaptation of feature
  detectors.
\newblock \emph{arXiv preprint arXiv:1207.0580}.

\bibitem[{{Hucko} and {Lacko}(2018)}]{hucko2018Diacritic}
A.~{Hucko} and P.~{Lacko}. 2018.
\newblock Diacritics restoration using deep neural networks.
\newblock In \emph{2018 World Symposium on Digital Intelligence for Systems and
  Machines (DISA)}, pages 195--200.

\bibitem[{Luu and Yamamoto(2012)}]{luu2012pointwise}
Tuan~Anh Luu and Kazuhide Yamamoto. 2012.
\newblock A pointwise approach for vietnamese diacritics restoration.
\newblock In \emph{2012 International Conference on Asian Language Processing},
  pages 189--192. IEEE.

\bibitem[{Maamouri et~al.(2004)Maamouri, Bies, Buckwalter, and
  Mekki}]{maamouri2004atb3}
Mohammed Maamouri, Ann Bies, Tim Buckwalter, and Wigdan Mekki. 2004.
\newblock The penn arabic treebank: building a large-scale annotated arabic
  corpus.
\newblock In \emph{NEMLAR Conference on Arabic Language Resources and Tools},
  pages 102–--109.

\bibitem[{Marton et~al.(2010)Marton, Habash, and Rambow}]{marton2010improving}
Yuval Marton, Nizar Habash, and Owen Rambow. 2010.
\newblock Improving arabic dependency parsing with lexical and inflectional
  morphological features.
\newblock In \emph{Proceedings of the NAACL HLT 2010 First Workshop on
  Statistical Parsing of Morphologically-Rich Languages}, pages 13--21.
  Association for Computational Linguistics.

\bibitem[{Mihalcea(2002)}]{mihalcea2002diacritics}
Rada~F Mihalcea. 2002.
\newblock Diacritics restoration: Learning from letters versus learning from
  words.
\newblock In \emph{International Conference on Intelligent Text Processing and
  Computational Linguistics}, pages 339--348. Springer.

\bibitem[{Mohamed Ould Abdallahi~Ould et~al.(2011)Mohamed Ould Abdallahi~Ould,
  Meziane, Mazroui, and Lakhouaja}]{bebah2011alkhalil}
Bebah Mohamed Ould Abdallahi~Ould, Abdelouafi Meziane, Azzeddine Mazroui, and
  Abdelhak Lakhouaja. 2011.
\newblock Alkhalil morphosys.
\newblock In \emph{7th International Computing Conference in Arabic}, pages
  66--73, Riyadh, Saudi Arabia.

\bibitem[{Mubarak et~al.(2019)Mubarak, Abdelali, Sajjad, Samih, and
  Darwish}]{mubarak2019highly}
Hamdy Mubarak, Ahmed Abdelali, Hassan Sajjad, Younes Samih, and Kareem Darwish.
  2019.
\newblock Highly effective {A}rabic diacritization using sequence to sequence
  modeling.
\newblock In \emph{Proceedings of the 2019 Conference of the North {A}merican
  Chapter of the Association for Computational Linguistics: Human Language
  Technologies, Volume 1 (Long and Short Papers)}, pages 2390--2395,
  Minneapolis, Minnesota. Association for Computational Linguistics.

\bibitem[{Nelken and Shieber(2005)}]{nelken2005arabic}
Rani Nelken and Stuart~M Shieber. 2005.
\newblock Arabic diacritization using weighted finite-state transducers.
\newblock In \emph{Proceedings of the ACL Workshop on Computational Approaches
  to Semitic Languages}, pages 79--86. Association for Computational
  Linguistics.

\bibitem[{Orife(2018)}]{orife2018attentive}
Iroro Orife. 2018.
\newblock Attentive sequence-to-sequence learning for diacritic restoration of
  yor$\backslash$ub$\backslash$'a language text.
\newblock \emph{arXiv preprint arXiv:1804.00832}.

\bibitem[{Osama~Hamed(2017)}]{HamedZesch:2017}
Torsten~Zesch Osama~Hamed. 2017.
\newblock {A Survey and Comparative Study of Arabic Diacritization Tools}.
\newblock \emph{JLCL}, 32(1):27--47.

\bibitem[{Pasha et~al.(2014)Pasha, Al-Badrashiny, Diab, El~Kholy, Eskander,
  Habash, Pooleery, Rambow, and Roth}]{pasha2014madamira}
Arfath Pasha, Mohamed Al-Badrashiny, Mona Diab, Ahmed El~Kholy, Ramy Eskander,
  Nizar Habash, Manoj Pooleery, Owen Rambow, and Ryan~M Roth. 2014.
\newblock Madamira: A fast, comprehensive tool for morphological analysis and
  disambiguation of arabic.
\newblock In \emph{LREC-2014}, Reykjavik, Iceland.

\bibitem[{Rashwan et~al.(2015)Rashwan, Al~Sallab, Raafat, and
  Rafea}]{rashwan2015deep}
Mohsen Rashwan, Ahmad Al~Sallab, M.~Raafat, and Ahmed Rafea. 2015.
\newblock Deep learning framework with confused sub-set resolution architecture
  for automatic arabic diacritization.
\newblock In \emph{IEEE Transactions on Audio, Speech, and Language
  Processing}, pages 505--516.

\bibitem[{Roth et~al.(2008)Roth, Rambow, Habash, Diab, and
  Rudin}]{roth2008arabic}
Ryan Roth, Owen Rambow, Nizar Habash, Mona Diab, and Cynthia Rudin. 2008.
\newblock Arabic morphological tagging, diacritization, and lemmatization using
  lexeme models and feature ranking.
\newblock In \emph{Proceedings of the 46th Annual Meeting of the Association
  for Computational Linguistics on Human Language Technologies: Short Papers},
  pages 117--120. Association for Computational Linguistics.

\bibitem[{Said et~al.(2013)Said, El-Sharqwi, Chalabi, and
  Kamal}]{microsoft2013diac}
Ahmed Said, Mohamed El-Sharqwi, Achraf Chalabi, and Eslam Kamal. 2013.
\newblock A hybrid approach for arabic diacritization.
\newblock In \emph{Natural Language Processing and Information Systems}, pages
  53--64, Berlin, Heidelberg. Springer Berlin Heidelberg.

\bibitem[{{\v{S}}anti{\'c} et~al.(2009){\v{S}}anti{\'c}, {\v{S}}najder, and
  Ba{\v{s}}i{\'c}}]{vsantic2009automatic}
Nikola {\v{S}}anti{\'c}, Jan {\v{S}}najder, and Bojana~Dalbelo Ba{\v{s}}i{\'c}.
  2009.
\newblock Automatic diacritics restoration in croatian texts.
\newblock \emph{INFuture2009: Digital Resources and Knowledge Sharing}, pages
  309--318.

\bibitem[{Schuster and Paliwal(1997)}]{schuster1997bilstm}
Mike Schuster and Kuldip~K Paliwal. 1997.
\newblock Bidirectional recurrent neural networks.
\newblock \emph{IEEE Transactions on Signal Processing}, 45(11):2673--2681.

\bibitem[{Tufi{\c{s}} and Ceau{\c{s}}u(2008)}]{tufics2008diac+}
Dan Tufi{\c{s}} and Alexandru Ceau{\c{s}}u. 2008.
\newblock Diac+: A professional diacritics recovering system.
\newblock \emph{Proceedings of LREC 2008}.

\bibitem[{Vergyri and Kirchhoff(2004)}]{vergyri2004automatic}
Dimitra Vergyri and Katrin Kirchhoff. 2004.
\newblock Automatic diacritization of arabic for acoustic modeling in speech
  recognition.
\newblock In \emph{Proceedings of the workshop on computational approaches to
  Arabic script-based languages, COLING'04}, pages 66--73, Geneva, Switzerland.
  Association for Computational Linguistics.

\bibitem[{Zitouni et~al.(2006)Zitouni, Sorensen, and
  Sarikaya}]{zitouni2006maximum}
Imed Zitouni, Jeffrey~S Sorensen, and Ruhi Sarikaya. 2006.
\newblock Maximum entropy based restoration of arabic diacritics.
\newblock In \emph{Proceedings of the 21st International Conference on
  Computational Linguistics and the 44th annual meeting of the Association for
  Computational Linguistics}, pages 577--584. Association for Computational
  Linguistics.

\bibitem[{Zweigenbaum and Grabar(2002)}]{zweigenbaum2002restoring}
Pierre Zweigenbaum and Natalia Grabar. 2002.
\newblock Restoring accents in unknown biomedical words: application to the
  french mesh thesaurus.
\newblock \emph{International Journal of Medical Informatics},
  67(1-3):113--126.

\end{thebibliography}

\label{lastpage}

\end{document}